\documentclass{article}

\usepackage{PRIMEarxiv}

\usepackage[utf8]{inputenc} 
\usepackage[T1]{fontenc}    
\usepackage{hyperref}       
\usepackage{url}            
\usepackage{booktabs}       
\usepackage{amsfonts}       
\usepackage{nicefrac}       
\usepackage{microtype}      
\usepackage{lipsum}
\usepackage{fancyhdr}       
\usepackage{graphicx}       
\graphicspath{{media/}}     

\title{A Review and Analysis of a Parallel Approach for Decision Tree Learning from Large Data Streams
}

\author{
  Zeinab Shiralizadeh
  \texttt{\ z.shiralizade1371@gmail.com} \\
}

\begin{document}
\maketitle

\begin{abstract}
This work studies one of the parallel decision tree learning algorithms, pdsCART, designed for scalable and efficient data analysis. The method incorporates three core capabilities. First, it supports real-time learning from data streams, allowing trees to be constructed incrementally. Second, it enables parallel processing of high-volume streaming data, making it well-suited for large-scale applications. Third, the algorithm integrates seamlessly into the MapReduce framework, ensuring compatibility with distributed computing environments. In what follows, we present the algorithm’s key components along with results highlighting its performance and scalability.

\end{abstract}


\section{Introduction}
The term Big Data \cite{sagiroglu2013big, fan2014challenges} refers to the growing need to process and analyze vast, complex datasets that are generated at high speed and in massive volumes \cite{tsai2015big, elgendy2014big}. Common examples include data exchanged on social media platforms \cite{brandes2013social}, weather observations \cite{olaiya2012application}, urban mobility records, and large-scale financial or commercial transactions \cite{bose2001business}. As the scale and frequency of data generation have increased, so have the challenges associated with managing and extracting meaningful insights. Key applications in this space include modeling the evolution of social networks \cite{fard2023temporal, fard2025robustness} and developing predictive models based on meteorological trends.

Another critical domain is streaming data, often produced by sensor networks \cite{yin2020dynamic} including mobile phones, distributed sensors, and systems for detecting repeated transmissions.
To address these challenges, a wide range of machine learning approaches have been proposed, offering effective frameworks for both analysis and decision-making in complex, high-volume data environments \cite{mahesh2020machine, hasanzadeh2019two}.

In this study, we focus on practical applications relevant to the second and third domains previously discussed, particularly those related to commercial and industrial data usage \cite{fard2023machine}.
With mobile phones now serving as a central means of communication, the volume of data generated through mobile apps and digital platforms has surged. In conjunction with website activity, these systems produce extensive user data, much of which is collected and analyzed by private companies.
Data from mobile devices, including signals from sensors and GPS systems \cite{lin2014mining}, yield rich, high-dimensional information that encompasses a wide variety of user and environmental attributes.

To handle such large-scale data, the MapReduce framework \cite{dean2008mapreduce} has emerged as a widely adopted standard for parallel data processing across distributed computing environments.
The framework is built around two fundamental operations: the Map step, which organizes, filters, and structures raw data, and the Reduce step, which aggregates and summarizes the output for downstream analysis.

While the MapReduce framework is well-suited for large-scale data processing, its design introduces limitations when applied to algorithms that require direct or online (non-batch) execution.
Driven by real-world commercial needs, our focus turned to decision tree algorithms \cite{patel2018study} as a viable approach for scalable, interpretable learning. In this work, we talk about one of the available methods for deploying decision tree models within the MapReduce framework to handle massive data streams effectively \cite{song2015decision}.

The remainder of this paper is structured as follows:
We begin by reviewing existing decision tree algorithms for data mining.
Next, we examine their application in the context of streaming data analysis.
Finally, we detail the implementation of the parallel decision tree approach, accompanied by a performance evaluation.

\section{Decision Tree for Big Data}
Decision trees remain one of the most effective and widely adopted methods in machine learning.
Their appeal lies not only in their flexibility and strong predictive performance, but also in their ability to produce transparent, rule-based outputs \cite{rokach2005clustering} that are easily interpretable by humans \cite{lipton2018mythos}.
This interpretability makes them especially valuable in domains such as business intelligence and data-driven decision-making.

That said, decision tree algorithms are not without limitations.
Early implementations \cite{kochenderfer2022algorithms, fard2023machine} often struggled with memory inefficiencies, as they required repeated access to large training datasets throughout the tree-building process.
Furthermore, identifying optimal split points typically demands sorting numerical features, a step that can add substantial computational overhead.
In response to these challenges, researchers have proposed a variety of strategies aimed at improving scalability and memory efficiency.

\subsection{Literature}
One of the common techniques used in decision tree algorithms is pre-sorting feature values, as seen in approaches like SPRINT \cite{song2022design} and ScalParC \cite{joshi1998scalparc}.
An alternative to this is estimating feature distributions rather than explicitly sorting them, a method adopted in algorithms such as SPIES, pCLOUDS, and SPDT \cite{gunarathne2010mapreduce}, which leverage histogram-based structures. These methods often involve multiple passes over the data to construct local histograms.

While pre-sorting strategies tend to offer high precision, they are not always well-suited for streaming data, where constant flow and limited memory render such computations impractical.

\subsubsection{Parallel Decision Tree Algorithms} 
Four major forms of parallelization have been proposed: horizontal, vertical, task-based, and hybrid.
Horizontal parallelization splits the entire dataset into smaller subsets for concurrent processing.
Vertical parallelization divides the features across processors.
Task-based parallelism distributes tree nodes across different tasks for independent growth.
Finally, hybrid parallelization integrates all three methods for maximum flexibility.

For instance, in the early stages of tree construction, a hybrid model may combine horizontal and vertical parallelism, then apply task-based parallelism at the final stages. A well-known example of this is Google’s PLANET framework \cite{panda2009planet}, which utilizes horizontal parallelism in the upper levels of the tree and shifts to task-based parallelism at the leaves, once the data fits into memory.

In other studies, researchers have used horizontal parallelism to construct histograms, which then guide decision-making in level-wise tree construction.

This technique has also been employed in gradient boosting trees, such as GBRT \cite{yang2022novel} or GBDT \cite{zhang2020gbdt}, and even in random forest models, where SPDT constructs local histograms during the Map phase of a MapReduce pipeline, which are then combined during the Reduce phase into a global histogram to determine optimal splits.

That said, a limitation remains: with each new tree level, the full dataset must be reprocessed. This multiple-pass requirement makes these methods unsuitable for true data stream scenarios.

\subsubsection{Decision Trees for Online Stream Mining}
Most traditional decision tree algorithms are built for batch processing and assume that data can be loaded into memory. However, with the rapid increase in data volumes, new methods are needed, ones capable of processing streaming data continuously without compromising valuable information.

In a streaming context, data arrives indefinitely, and the model must decide on the best splitting feature using only a single pass through the data. This need has led to the development of specialized stream-friendly methods, such as the Hoeffding Tree algorithm and the Very Fast Decision Tree (VFDT) \cite{jia2020vfdt}, which apply Hoeffding’s inequality to ensure statistically sound splits with limited data.

Advanced versions like CVFDT extend this idea with vertical parallelism, enabling scalable online classification. These methods are all built on the statistical foundation of Hoeffding’s bound.

However, since Hoeffding’s inequality does not always capture the full probabilistic complexity of streaming data, newer approaches have been proposed. For example, adaptations of traditional decision tree algorithms like ID3, C4.5, and CART have been reimagined for streaming data. One such model, dsCART \cite{rutkowski2014cart}, uses Gaussian estimation to identify the most suitable feature for splitting a node.

A key insight from this work is that the feature selected by dsCART with high probability matches the feature that would have been selected by analyzing the entire dataset in hindsight, but it does so in a single pass.

\subsection{A Parallel Approach for Data Streams}
In this work, we do our study based on a parallel approach, pdsCART, designed to build decision trees for large-scale data stream classification and prediction.
As the foundation for our study, we selected dsCART, a decision tree algorithm originally developed for stream-based classification tasks.
Our solution adapts dsCART to support horizontal parallelization, implemented within the MapReduce programming model.

Although several prior efforts such as PLANET, SPDT, SRF, and GBDT have leveraged MapReduce to scale various tree-based algorithms, these methods were not specifically developed for single-path decision trees in streaming environments. Notably, none of these approaches have been tailored to support stream-based learning in the context of directional decision trees like GBRT.

The implementation details of this method are discussed further in the next section.

\section{PdsCART Implementation}
In this section, we describe the discussed method, which focuses on parallelizing the decision tree algorithm.
To that end, we begin by explaining both PdsCART, the parallel approach, and the original dsCART algorithm, which was designed for stream data classification.
We then outline the implementation details of PdsCART within the MapReduce framework.

It is important to emphasize that our objective is not to introduce an entirely new model, but rather to demonstrate that existing learning models can be adapted for parallel execution, thereby significantly reducing training time.

\subsection{Preliminary Considerations}
In the CART algorithm, identifying the optimal split point tends to be the most time-consuming operation, particularly when using the Gini index. For each feature at a given node, the algorithm must evaluate all possible partitions of the feature’s value set to determine the best improvement. These computations must be performed for every new instance received from the data stream, which consumes both time and memory.

Interestingly, the feature selected at a given node based on the current stream segment often closely aligns with the one that would have been chosen had the entire dataset been available. This means the exact timing of the estimation is less important, as the probability of selecting the correct feature remains consistent. This insight motivates our approach to estimating and controlling split conditions after processing a variable number of instances independently.

By selecting an appropriate high-confidence threshold, our algorithm is able to produce decision trees nearly identical to those generated by dsCART, but with significantly faster processing times.
The parallelization strategy of the PdsCART algorithm is outlined in the next subsection, and its experimental results are presented in the evaluation section.

\subsection{MapReduce Implementation}
In the context of decision trees, PdsCART adopts a horizontal partitioning strategy implemented within the distributed MapReduce framework.
The controller orchestrates the tree's growth, while the Map and Reduce tasks perform their standard responsibilities.

Suppose we have 
R total records and want to divide them across 
P parallel mappers. The controller distributes approximately
R/P records to each mapper for processing.
To support split-point decisions, PdsCART uses simple, repeatable data structures, namely, histograms, to track the frequency of features and class labels. These local histograms are used to compute the Gini index improvements for candidate splits at each leaf node.

During the Map phase, each incoming record is routed to its corresponding leaf node, where the mapper updates its local histogram accordingly.
In the Reduce phase, reducers collect the local histograms from all relevant mappers and merge them into a single global histogram for each leaf. The results are written to a structured output file.

Using these merged histograms, the controller can traverse the global histograms across all leaves, estimate the best splitting features, and decide whether to grow the tree further based on the current split criteria.

Each mapper receives the current version of the tree along with a subset of records. For every record, it traverses the tree to locate the target leaf node.
Based on the record’s ID and leaf assignment, the mapper updates the corresponding local histogram with the new data.

Once all records for a given round are processed, the mapper clears the histograms and node assignments tied to those record IDs.
During the Reduce phase, all local histograms are aggregated to form the global statistics for each node. These are then written to output files for the controller to use in the next round of tree growth.

With a single pass over the output file, the controller can now compute the top two candidate splitting features for each leaf node in the tree. The PdsCART algorithm then proceeds to evaluate the splitting conditions, based on a specified threshold parameter $\Theta$ that governs node splitting behavior. If needed, the algorithm will split the corresponding leaf nodes accordingly.

In the experimental section, we demonstrate that in our proposed approach, feature selection estimations are computed infrequently, only after a sufficiently large number of samples have been processed, and are executed in parallel.
Despite this relaxed evaluation frequency, the resulting decision trees are comparable to those produced by dsCART, while requiring less overall computation time due to more efficient, distributed processing.

\section{Experiments}
This section summarizes several key findings from our experimental evaluation.
The proposed parallel learning approach was specifically designed to support decision tree induction from data streams.
We found that, with appropriate tuning of the parameters, our method consistently produced results that matched those of dsCART. In fact, across all of our test cases, aside from the number of records processed, we observed the exact same decision trees being generated with both implementations, and with identical accuracy levels.

Given this outcome, there is no need to compare learning models based on predictive performance.
Instead, our focus shifts toward evaluating and highlighting the theoretical and practical efficiency gains achieved through our parallel implementation.

To that end, we begin by describing the experimental scenarios and the characteristics of the datasets used, and then present the results obtained from implementing our proposed method.

\subsection{Experimental Scenarios}
Processing larger volumes of data in parallel naturally reduces overall execution time, but we also considered several additional factors to comprehensively evaluate the performance of our parallel approach.
Beyond execution time, the aspects under investigation included:
the number of records per iteration,
the number of features,
buffer size, and
the split confidence parameter.

Other potential metrics, such as the size of the resulting decision tree or interdependencies among variables, are noted for consideration in future work.

Importantly, these parameters do not operate independently.
To study their relationships, we designed five synthetic data stream configurations along with one real-world dataset derived from the KDD CUP 99 collection, obtained through MOA.
While the synthetic datasets are generated to test a range of controlled combinations, the KDD dataset serves as a realistic benchmark for validating the dsCART algorithm.
The characteristics of all related datasets are summarized below in Table 1. 

\begin{table}[h!]
\label{tab:datasets}
\centering
\begin{tabular}{|c|c|c|c|c|c|}
\hline
\# & \textbf{Data Set} & \# \textbf{records} & \textbf{Attributes} & \textbf{Classes} & \# \textbf{type} \\
\hline
1 & $D_1$ & 10 thousands & 5 & 2 & synthetic \\
2 & $D_2$ & 500 thousands & 70 & 5 & synthetic \\
3 & $D_3$ & 1.5 millions & 20 & 10 & synthetic \\
4 & $D_4$ & 4 millions & 10 & 5 & synthetic \\
5 & $D_5$ & 4 millions & 15 & 2 & synthetic \\
6 & $D^*$ & 4.8 millions & 34 & 20 & web data \\
\hline
\end{tabular}
\caption{Overview of data sets}
\end{table}

\subsection{Experimental Results}
The initial results, as shown in Table 2, with PdsCART show that it can produce decision trees that are both accurate and structurally identical to those generated by dsCART, while achieving notable improvements in execution time.
This comparison highlights some of the most important outcomes that led to the development of the PdsCART algorithm.

\begin{table}[h!]\label{tab:record_accuracy_time}
\centering
\begin{tabular}{|c c c|c c c|c c c|}
\hline
\multicolumn{3}{|c|}{$D_1$} & \multicolumn{3}{c|}{$D_5$} & \multicolumn{3}{c|}{$D^*$} \\
\hline
Records & Accuracy & Time & Records & Accuracy & Time & Records & Accuracy & Time \\
\hline
1   & 83.11\% & 11.44  & 1   & 84.94\% & 5076.77 & 1   & 77\%    & 9106.35 \\
20  & 83.11\% & 2.10   & 200 & 84.94\% & 318.68  & 200 & 77\%    & 409.98  \\
40  & 83.11\% & 1.56   & 400 & 84.94\% & 253.71  & 400 & 77\%    & 297.19  \\
60  & 83.11\% & 1.35   & 600 & 84.94\% & 215.75  & 600 & 77\%    & 250.04  \\
80  & 83.11\% & 1.23   & 800 & 84.94\% & 183.67  & 800 & 77\%    & 227.55  \\
\hline
\end{tabular}
\caption{Accuracy and Time for Varying Record Counts Across Data Sets}
\label{tab:record_accuracy_time}
\end{table}

Looking specifically at the accuracy column, we observe that even as the number of instances used for computing split points increases, the model's accuracy remains consistently stable.
This further confirms the reliability of our method in maintaining performance while benefiting from parallelized processing.

This specific characteristic is crucial for understanding the behavior of the parallel algorithm. Leveraging it, the proposed method can significantly reduce execution time, as shown in the time column of our results. These findings are further supported by additional experiments.

For instance, faster processing times are achieved when fewer computations are needed to determine split conditions.
Take dataset D4 as an example: when processing 4 million records in batches of 22, around 20,000 computations are required. But when processing batches of 800 records, only about 5,000 computations are necessary.
This suggests that for real-time commercial applications, where predictive models must react quickly to changes in the data stream, a smaller number of records per split may be more suitable.
On the other hand, for use cases that can tolerate delay or require broader aggregation, larger batches may lead to more efficient processing.

Additionally, early discovery of promising split points can simplify future calculations. New leaf nodes will then receive fewer records for their own split estimations, which helps explain why processing time doesn’t always increase linearly with larger record counts, in fact, it's sometimes lower than with smaller batches.

For example, when comparing datasets D4 and D5, both containing the same number of records, we observe that execution time is significantly better for D5 under similar test configurations.
This difference is due to the number of features: datasets with more features naturally require more time to process, as shown in Table 3.
The results for all four synthetic datasets confirm this pattern—each contains 4 million records, but varies in feature count.

\begin{table}[h!]\label{tab:tree_stats}
\centering
\begin{tabular}{|c|c|c|c|c|c|c|}
\hline
\# & \textbf{Data Set} & \textbf{TreeDepth} & \textbf{TreeNodes} & \textbf{TreeLeaves} & \# \textbf{Attributes} & \textbf{Time} \\
\hline
1 & $D_{2a}$   & 2  & 3    & 2    & 2  & 19.59   \\
2 & $D_{5a}$   & 5  & 29   & 15   & 5  & 66.54   \\
3 & $D_{10a}$  & 10 & 213  & 107  & 10 & 259.651 \\
4 & $D_{20a}$  & 20 & 5501 & 2197 & 20 & 1610.75 \\
\hline
\end{tabular}
\caption{Execution time on datasets with different numbers of attributes}
\label{tab:tree_stats}
\end{table}

Another notable result concerns the computation time required for evaluating splits.
When PdsCART needs to select the best splitting feature, the algorithm must examine all possible partitions of each feature’s value set. This process is directly influenced by the number of bins used in the histograms.

As expected, histograms, shown in Table 4, with more bins require more time to evaluate all potential partitions, since the algorithm must assess a finer-grained distribution of values.
This characteristic also opens the door for a new level of parallelization, where each partition could potentially be analyzed independently.

\begin{table}[h!]\label{tab:bin_results}
\centering
\begin{tabular}{|c|c|c|c|c|c|c|}
\hline
\# & \textbf{Data Set} & \textbf{2 bins} & \textbf{4 bins} & \textbf{6 bins} & \textbf{8 bins} & \textbf{10 bins} \\
\hline
1 & $D_2$ & 73.28 & 77.41 & 82.36 & 89.40 & 98.63 \\
2 & $D_3$ & 79.96 & 83.70 & 90.97 & 100.22 & 111.77 \\
3 & $D_4$ & 145.55 & 153.22 & 167.28 & 204.38 & 229.88 \\
4 & $D_5$ & 151.89 & 166.16 & 190.98 & 250.27 & 313.59 \\
\hline
\end{tabular}
\caption{Performance across varying bin sizes}
\label{tab:bin_results}
\end{table}

We view this as a promising direction for future work, where additional layers of parallelism can be introduced to further accelerate the evaluation process.

\section{Conclusion and Future Work}
In this paper, we studied one of the methods for implementing a decision tree within the MapReduce framework.
A major advantage of the proposed algorithm is its ability to construct a decision tree in a single pass over the data. This is particularly valuable in the context of data streams, where multiple passes over the same dataset are often impractical or even impossible.

Another important strength of the method is its implementation efficiency. We showed that the algorithm performs exceptionally well when handling large volumes of data in parallel, achieving strong scalability.
These findings provide a solid foundation for future research.

One area of future work is to analyze how the algorithm scales with an increasing number of processing units, and to better understand how other parameters influence the algorithm’s performance.
While some of these effects are relatively intuitive, more in-depth evaluation is needed to assess their impact on tree quality.

We have already demonstrated that the algorithm can achieve error rates comparable to other decision tree learners, but other characteristics—such as tree size, depth, and feature ordering—also warrant closer investigation in future studies.

\bibliographystyle{unsrt}  
\bibliography{references}

\end{document}